\documentclass{article}
\usepackage{spconf,amsmath,epsfig}

\usepackage{subfigure}
\usepackage{array}
\usepackage{framed,multirow}
\newcommand{\tabincell}[2]{\begin{tabular}{@{}#1@{}}#2\end{tabular}}
\begin{document}\sloppy

\def\x{{\mathbf x}}
\def\L{{\cal L}}

\title{BOUNDARY AWARE MULTI-FOCUS IMAGE FUSION USING DEEP NEURAL NETWORK}
%
\name{Haoyu Ma, Juncheng Zhang, Shaojun Liu, Qingmin Liao}
\address{Shenzhen Key Laboratory of Information Science and Technology, \\
Shenzhen Engineering Laboratory of IS\&DRM, Department of Electronics Engineering, \\
Graduate School at Shenzhen, Tsinghua University, Shenzhen, 518055, China \\
\{hy-ma17, zjc16, liusj14\}@mails.tsinghua.edu.cn; liaoqm@tsinghua.edu.cn}

\maketitle

\begin{abstract}
Since it is usually difficult to capture an all-in-focus image of a 3D scene directly, various multi-focus image fusion methods are employed to generate it from several images focusing at different depths. However, the performance of existing methods is barely satisfactory and often degrades for areas near the focused/defocused boundary (FDB). In this paper, a boundary aware method using deep neural network is proposed to overcome this problem. (1) Aiming to acquire improved fusion images, a 2-channel deep network is proposed to better extract the relative defocus information of the two source images. (2) After analyzing the different situations for patches far away from and near the FDB, we use two networks to handle them respectively. (3) To simulate the reality more precisely, a new approach of dataset generation is designed. Experiments demonstrate that the proposed method outperforms the state-of-the-art methods, both qualitatively and quantitatively.
\end{abstract}
%

\section{Introduction}

When capturing an image of a 3D scene, it is difficult to take an image where all the objects are focused since the depth-of-field is limited. However, in many image processing tasks, it is more convenient and effective to use all-in-focus images as the input. Multi-focus image fusion, a technic to generate an all-in-focus image from several images of the same scene focusing on different depths, is effective to address this problem. For decades, a large number of multi-focus image fusion algorithms have been proposed. Most of them can be broadly categorized into two groups, i.e., transform domain-based algorithms [1-7] and spatial domain-based algorithms [8-15].

The transform domain-based algorithms are usually based on multi-scale transform (MST) theories, such as the Laplacian pyramid (LP) \cite{RN33}, wavelet transform \cite{RN30,RN72}, curvelet transform (CVT) \cite{RN31}, and non-subsampled contourlet transform (NSCT) \cite{ZHANG20091334}. These methods usually decompose the source images first, then extract and fuse the features of the images, and finally reconstruct the fused images. In addition to MST methods, some feature space-based methods have been proposed in recent years, such as independent component analysis (ICA) \cite{RN28}, sparse representation (SR) \cite{RN35} and NSCT-SR \cite{LIU2015147}. Most of the transform domain methods are efficient, but the fusion images are usually indistinct.

The spatial domain-based algorithms can be further divided into block-based, region-based and pixel-based fusion algorithms. The block-based algorithms usually divide an image into blocks, measure their spatial frequency as well as sum modified-Laplacian \cite{RN39}, and then fuse the image blocks. In these algorithms, such as morphology-based fusion \cite{RN27}, the size of the image block has a great impact and is hard to decide. The region-based algorithms \cite{RN85} are based on segmentation of input images and usually highly depend on segmentation accuracy. Recently, several pixel-based algorithms, including guided filtering (GF) \cite{RN80} and dense SIFT (DSIFT) \cite{RN43}, have been proposed. They can achieve improved results, however, usually suffer from the blocking effect.

\begin{figure*}[!ht]
\centering 
   \includegraphics[width=0.80\linewidth]{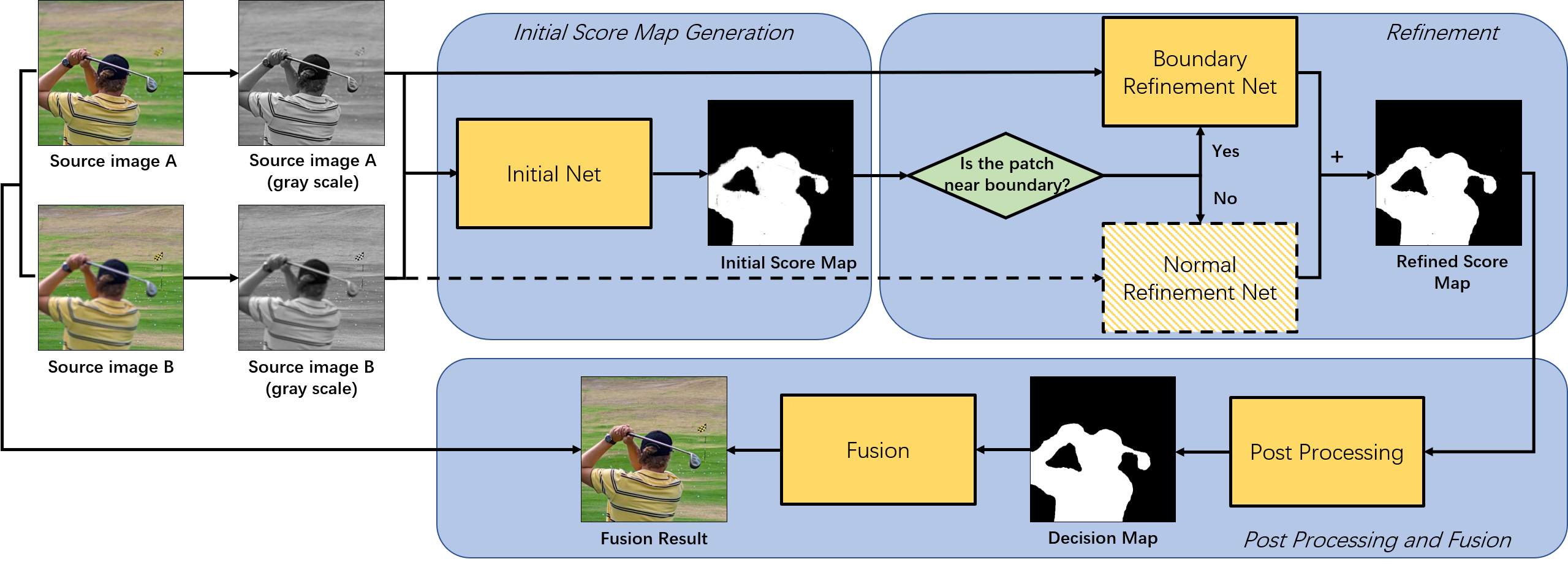}
   \caption{The block diagram of the proposed boundary aware multi-focus fusion method.}
\label{fig:flowchart}
\end{figure*}

For all of these traditional fusion methods, both transform domain-based and spatial domain-based, the defocus level descriptors and the comparison rules need to be designed manually. However, the situations are quite complex for real photos, and the results of these methods are usually imperfect. In \cite{RN497}, CNN is used to extract the defocus level descriptors and the comparison rules in a data-driven way. Unfortunately, the results of existing neural network based approaches \cite{RN497,RN91} are still unsatisfied, especially for the areas near the focused/defocused boundary (FDB), therefore much post processing is employed to ease this problem. The reasons are explained as follows. Firstly, the network structures used in \cite{RN497,RN91} could be improved for the fusion task. Secondly, the situations for areas far away from and near the FDB are quite different, therefore, it is unwise using a single network to tackle these two situations together. Thirdly, training datasets used by  \cite{RN497,RN91} are inconsistent with the reality.

In this paper, a boundary aware multi-focus image fusion approach using deep neural network is proposed to overcome the insufficiency, together with a new dataset generation method. The contributions are threefold. (1) Compared with existing networks for fusion \cite{RN497,RN91}, a 2-channel structure is designed to better extract the relative defocus information of the two source images and improve the fusion results. (2) Instead of a single network, two networks are employed to tackle the different situations for areas far away from and near the FDB respectively. As a consequence, the FDB of the fusion image is more clear for the proposed method. (3) To simulate the reality better, especially the situation for area near the FDB, a more reasonable approach of dataset generation is presented. Based on these three improvements, the proposed method can obtain pleased fusion results. Experiments demonstrate that the proposed method outperforms the state-of-the-art methods, both qualitatively and quantitatively.

\section{Proposed Fusion Method}

Our method consists of 3 steps: initial score map generation, score map refinement, as well as post processing and image fusion. The block diagram of our method is shown in Fig. \ref{fig:flowchart}. Firstly, a 2-channel network is proposed to generate an initial score map. Secondly, the initial score map is refined by two other networks that are specially designed for areas near and far away from the FDB. Thirdly, some simple post processing is employed to generate the decision map from the refined score map. Then the fusion result is obtained from the two source images according to the decision map.


\subsection{Initial Score Map Generation}

Pixel-wised multi-focus image fusion can be viewed as a classification problem, and CNN is the classic model for image classification. Therefore it is natural to begin with a simple CNN. In order to get better fusion results, a network deeper than \cite{RN497,RN91} is needed. However, when a network is deepened, the degradation problem would arise, because the optimization is not similarly easy for all systems. The residual learning framework fixes this problem and achieves precise classification \cite{RN618}. Therefore, it is employed here.

\begin{figure}[!ht]
\centering 
  \includegraphics[width=0.80\linewidth]{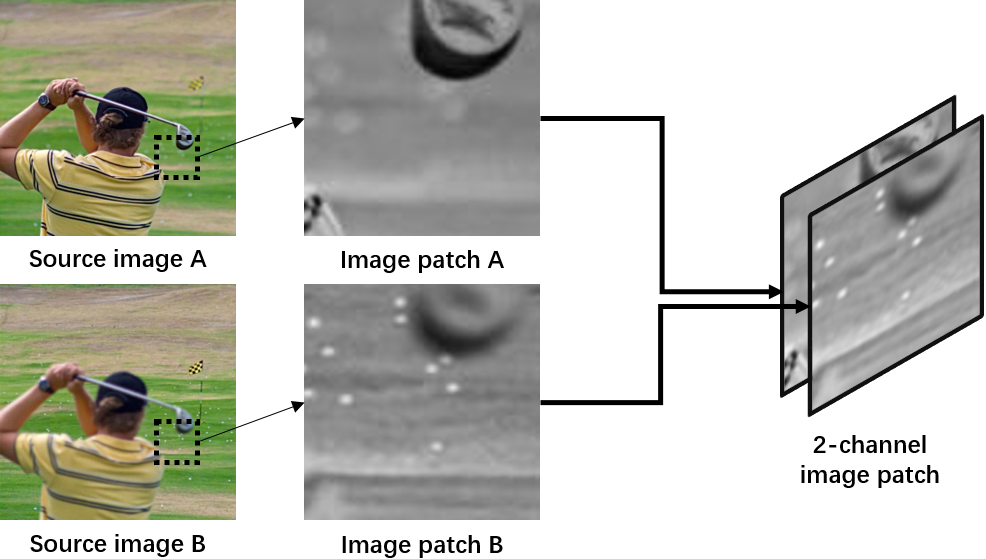}
  \caption{The input patch of the 2-channel network.}
\label{fig:subfig:channel2} 
\end{figure}

Since only the comparative information of input images is desired for multi-focus image fusion task, variety of structures might be applied. Tang \cite{RN91} used a single input network, which paid little attention to the relativity between the source images. Using the two source images together as input should be a better choice. Furthermore, for multi-focus image fusion, two source images are taken from the same position, meaning that they are perfectly aligned in nature. The same result is expected to gain when the order of two source images is changed. Therefore, for pixel-wise image fusion, the 2-channel model is more effective and flexible than the siamese model \cite{RN619}. As it's demonstrated in Fig. \ref{fig:subfig:channel2}, the two input source image patches are simply taken as one image patch with two channels.

\subsection{Refinement}

The area near the FDB of the fusion results is unsatisfied even using the proposed network. There are two reasons.

Firstly, the situations are quite different for patches far away from and near the FDB. For the patches far away from the FDB, the patches are totally focused or defocused. Consequently, the sharpness of the patch can be used as a metric to generate the score map. On the contrary, for patches near the FDB, both focused area and defocused area exist. Therefore, the sharpness of the patch is not suitable anymore, and the position of the center pixel becomes even more vital, since the proposed method is pixel-wise. For example, the patch should be thought as focused if the center pixel is in the focused region, even if the defocused area is larger in the whole patch. In conclusion, it is unwise to handle these two different situations using a single network. 

Secondly, the number of patches near the FDB is much fewer than that far away from the FDB, as the training patches are usually chosen from training images randomly. The imbalance of the training patches will also lead to the bad performance near the FDB when using a single network.

Therefore, taking the defocus level comparison task apart as two relatively independent one might be a better choice. Specifically, based on the initial score map, each patch can be classified to either far away from or near the FDB. Then they are processed by the normal refinement net and the boundary refinement net, respectively, to obtain a better score map. It is worth to point out that the initial net already has a satisfied performance over the patches far away from the boundary. Therefore, the initial fusion net can serve as the normal refinement net directly without any loss in performance. Ultimately, the flow path of the proposed method is simplified. After the initial score map generation, the patches near the FDB are processed by the boundary refinement net. Then the initial focus scores of these position are replaced.

The classification of patches near the FDB is described as follows. Based on the initial score map, the average of focus score ($FS$) at the centre pixel $(m,n)$ of a patch is calculated in a surrounding window of size $(2l+1) \times (2l+1)$ as follows.
\begin{eqnarray}
\overline{FS(m,n)}= \frac{1}{(2l+1)^2}\sum\nolimits _{i=m-l}^{i=m+l}\sum\nolimits _{j=n-l}^{j=n+l}FS(i,j).
\end{eqnarray}
If the average focus score satisfies 
\begin{eqnarray}
0.2 < \overline{FS(m,n)} < 0.8, 
\end{eqnarray}
the patch is taken as near the FDB.

\subsection{Post Processing and Image Fusion}

After the refinement, some simple post processing such as binarization and small region removal \cite{RN497} are still needed. Specifically, binarization is applied to the focus score map to obtain a decision map. Small regions inside the focused or defocused area are also cleared. These small tricks contribute some improvement to the final result, making it more visually comfortable. We intend to finish the main job by the networks, therefore, more post processing used in existing works, such as guided filter \cite{RN497}, is not needed.

Once the decision map is obtained, the fusion image ($ImgF$) can be directly generated from the decision map ($DM$) and the two source images ($ImgS$) as follows.
\begin{eqnarray}
ImgF=DM \cdot ImgS_A+(1-DM) \cdot ImgS_B.
\end{eqnarray}

\begin{figure}[!ht]
\centering 
  \subfigure[Source image A]{\label{fig:subfig:e21}
    \includegraphics[width=0.45\linewidth]{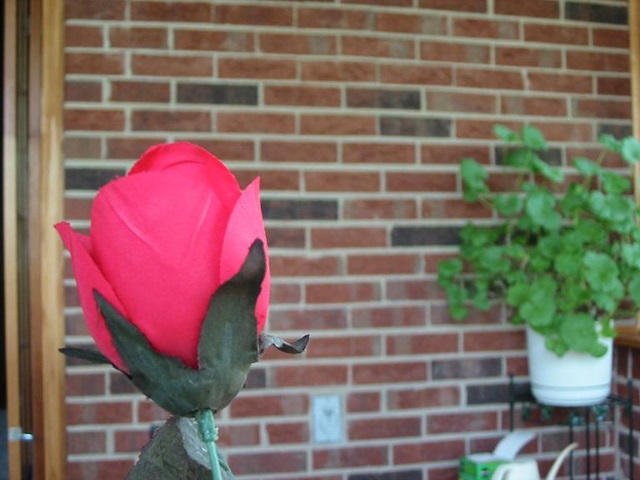}}
  \subfigure[Source image B]{\label{fig:subfig:e22}
    \includegraphics[width=0.45\linewidth]{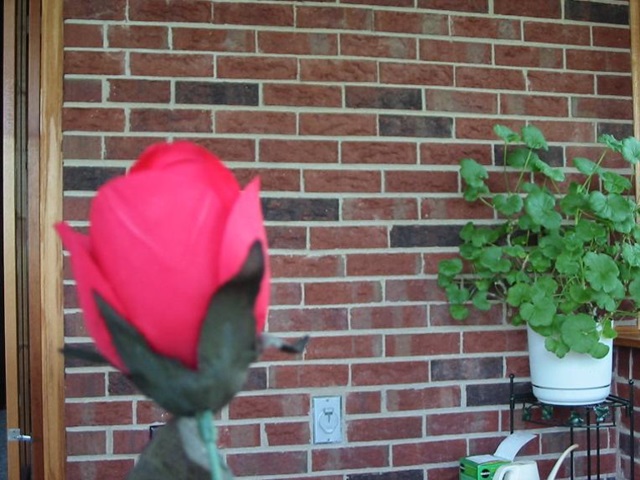}}
  \subfigure[After Initial Net]{\label{fig:subfig:e23}
    \includegraphics[width=0.45\linewidth]{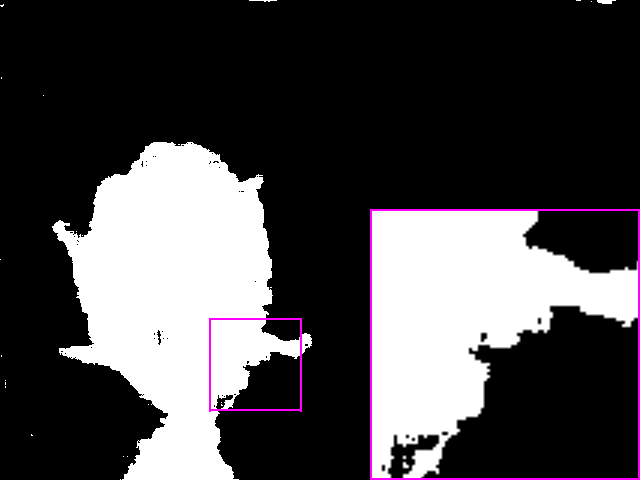}}
  \subfigure[After Refinement]{\label{fig:subfig:e24}
    \includegraphics[width=0.45\linewidth]{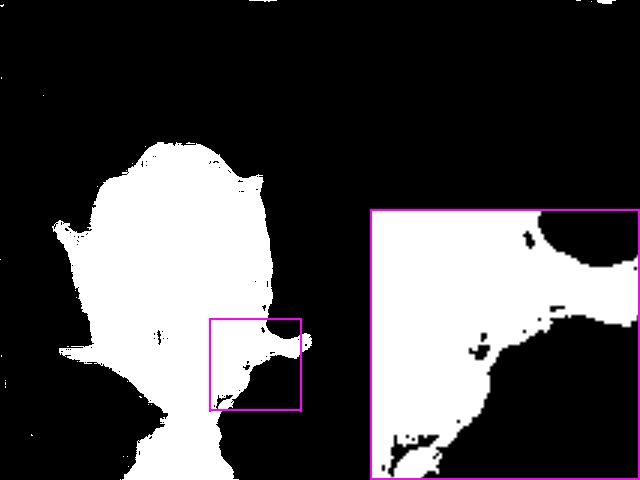}}
  \subfigure[Decision Map]{\label{fig:subfig:e25}
    \includegraphics[width=0.45\linewidth]{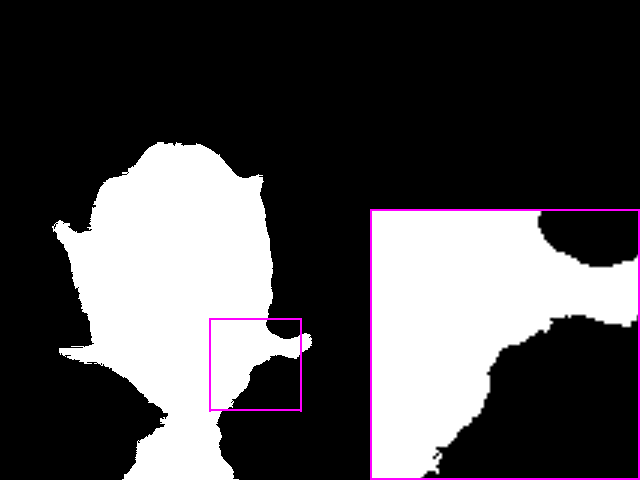}}
  \subfigure[Fusion Result]{\label{fig:subfig:e26}
    \includegraphics[width=0.45\linewidth]{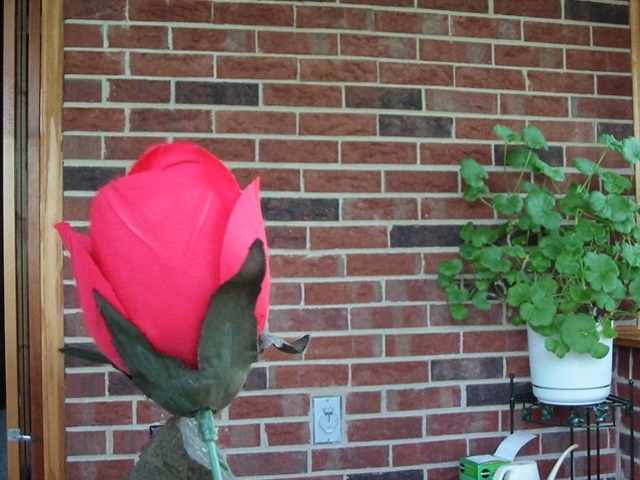}}
  \caption{The fusion maps after each step and the fusion result.}
\label{fig:subfig:7} 
\end{figure}

The score maps after each step are shown in Fig. \ref{fig:subfig:7}. As shown in Fig. \ref{fig:subfig:e23}, the score map after the initial fusion net is decent, but imperfections exist near the FDB. Fig. \ref{fig:subfig:e24} shows that the refinement net fixes these defects, and the result near the FDB becomes better. Then the post processing removes the small regions inside, and the decision map is given in Fig. \ref{fig:subfig:e25}. The fusion image is shown in Fig. \ref{fig:subfig:e26}.

\section{Experiments}

In this section, firstly a new approach to generate training dataset for deep-learning based multi-focus image fusion is proposed. Then the network details and the settings of training are discussed. Next, the comparative experiments with existing approaches are set. Finally, the results of proposed fusion method are shown and discussed.

\subsection{Proposed Dataset Generation Method }

A good training dataset should represent the normal and comprehensive situations of the task. In other fusion methods based on machine learning \cite{RN497,RN91}, several training images generation approaches have been used. Unfortunately, none of them simulates the reality that the FDB usually coincides with the object boundary \cite{7589980}.

Considering the reality, the best choice is to use real photos. However there are only few multi-focus fusion source images, and the ground truth needs to be labeled manually. Therefore, a feasible method is to generate artificial training images that are similar to the reality yet easy to obtain. The foreground images dataset with ground truth is used and some images without obvious defocus are chosen as the background dataset. Both the original foreground ($FG$) and the background ($BG$) images are processed by Gaussian filters for the blurred images firstly. Then the source image pairs are obtained according to the ground truth ($GT$) pixel by pixel.
\begin{eqnarray}
ImgS_A= FG_{Ori} \cdot GT+BG_{Blu} \cdot (1-GT), \\
ImgS_B= FG_{Blu} \cdot GT+BG_{Ori} \cdot (1-GT).
\end{eqnarray}

Specifically, Alpha Matting dataset \cite{AlphaMating}, which contains 27 images with ground truth, are used as foreground dataset, and 700 background images from COCO 2017 dataset \cite{COCO} are used as background dataset. The foreground and background images are combined one by one, therefore, 18,900 pairs of source image are obtained. For each source image pair, we randomly select 10 patches pairs. A training pair will be labeled 1 if the center pixel is focused in the source image A and defocused in the source image B, otherwise labeled 0. We then exchange the channels of a patch pair and label it the opposite, so the changing of input order will lead to the same result. Each source image pair is augmented by rotating and flipping. Finally, there are 2,268,000 training samples in total.

Additionally, the foreground images from Alpha Matting are larger than the background images from COCO dataset, so the foreground images are resized half at the very start.

\begin{figure*}[ht]
\centering 
  \subfigure[Source image A]{\label{fig:subfig:a11}
    \includegraphics[width=0.19\linewidth]{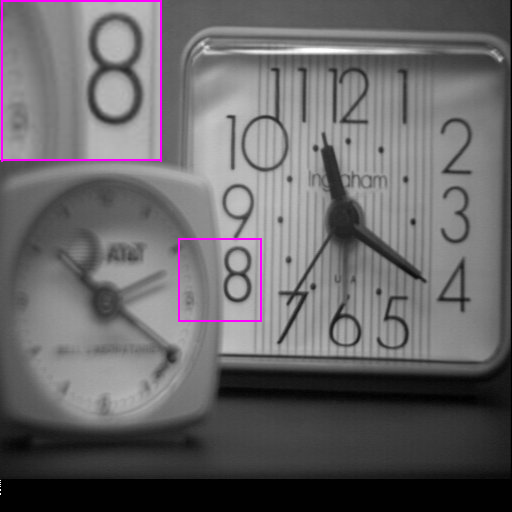}}
  \subfigure[Source image B]{\label{fig:subfig:b11}
    \includegraphics[width=0.19\linewidth]{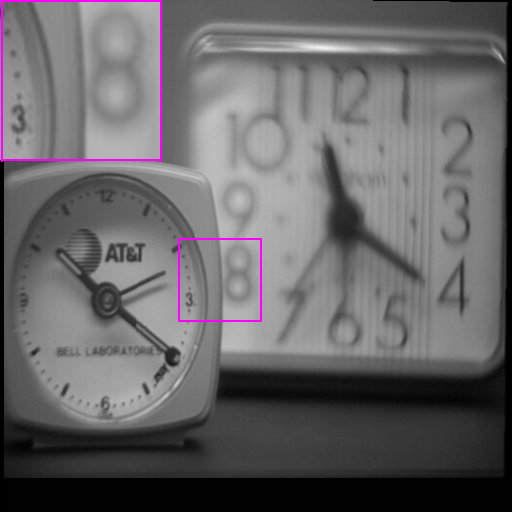}}
  \subfigure[NSCT]{\label{fig:subfig:c11}
    \includegraphics[width=0.19\linewidth]{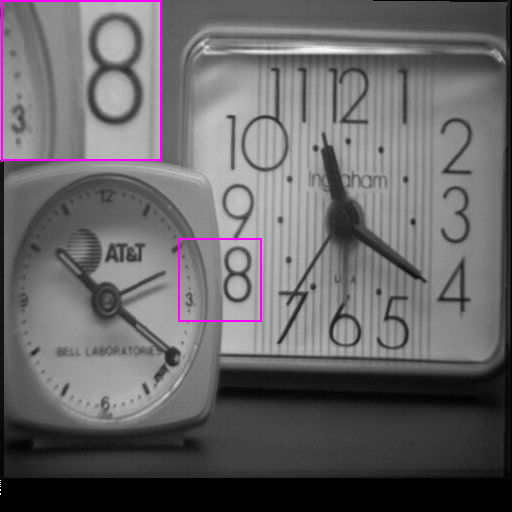}}
  \subfigure[SR]{\label{fig:subfig:d11}
    \includegraphics[width=0.19\linewidth]{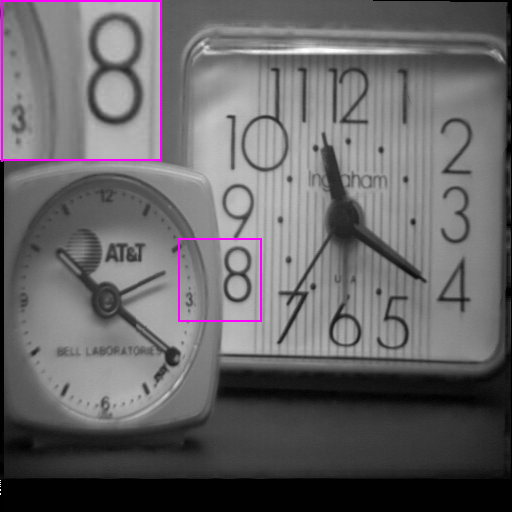}}\\
  \subfigure[NSCT-SR]{\label{fig:subfig:e11}
    \includegraphics[width=0.19\linewidth]{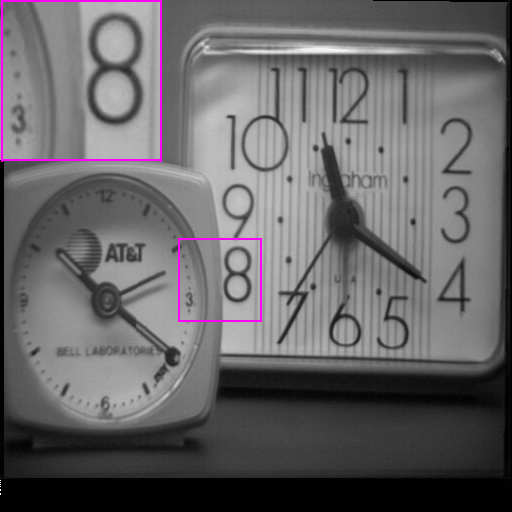}}
  \subfigure[GF]{\label{fig:subfig:f11}
    \includegraphics[width=0.19\linewidth]{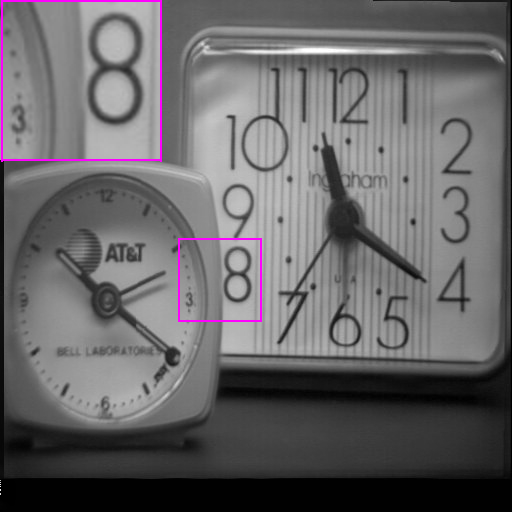}}
  \subfigure[DSIFT]{\label{fig:subfig:g11}
    \includegraphics[width=0.19\linewidth]{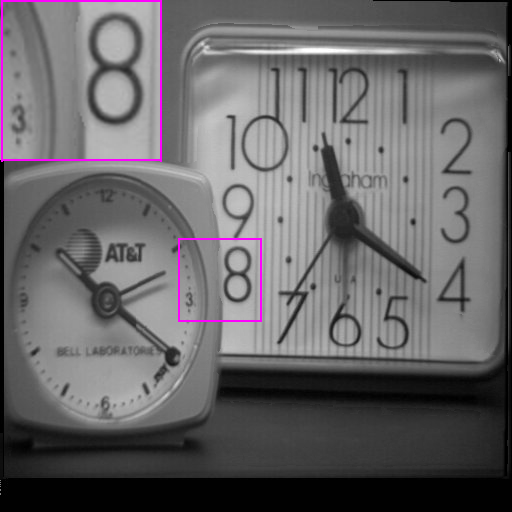}}
  \subfigure[CNN]{\label{fig:subfig:h11}
    \includegraphics[width=0.19\linewidth]{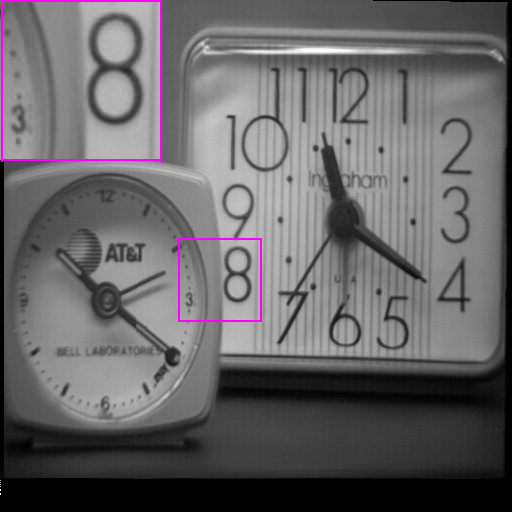}}
  \subfigure[The proposed method]{\label{fig:subfig:i11}
    \includegraphics[width=0.19\linewidth]{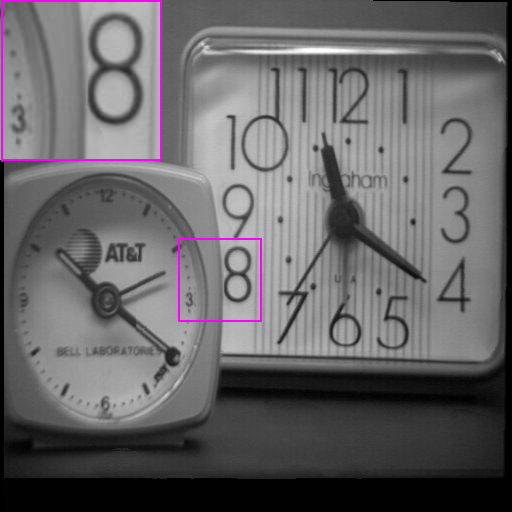}}
  \caption{The fusion results of different methods on the "clock".}
\label{fig:subfig:3} 
\end{figure*}

\begin{figure*}[!ht]
\centering 
  \subfigure[Source image A]{\label{fig:subfig:a12}
    \includegraphics[width=0.19\linewidth]{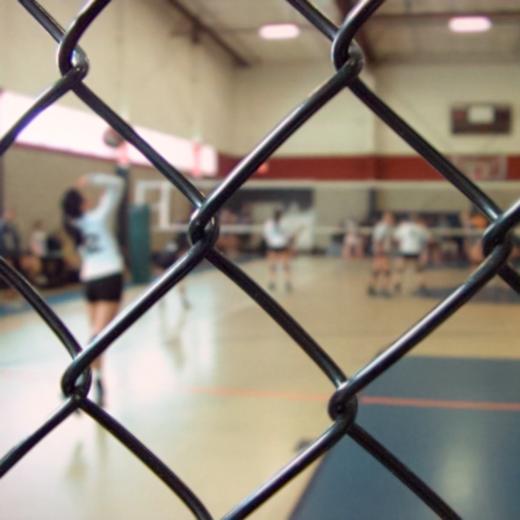}}
  \subfigure[Source image B]{\label{fig:subfig:b12}
    \includegraphics[width=0.19\linewidth]{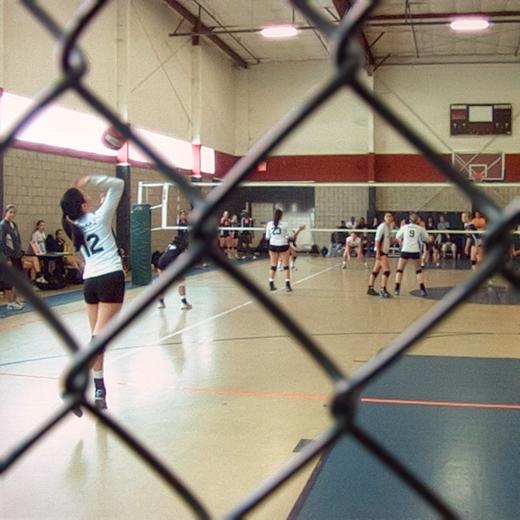}}
  \subfigure[NSCT]{\label{fig:subfig:c12}
    \includegraphics[width=0.19\linewidth]{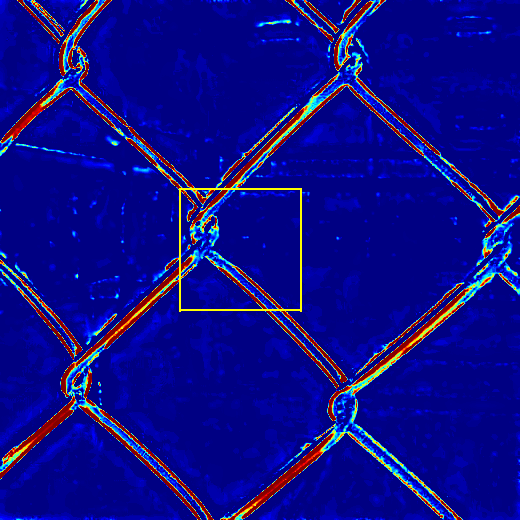}}
  \subfigure[SR]{\label{fig:subfig:d12}
    \includegraphics[width=0.19\linewidth]{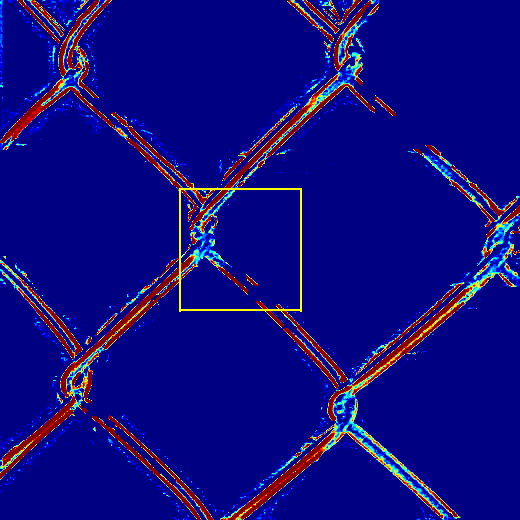}}\\
  \subfigure[NSCT-SR]{\label{fig:subfig:e12}
    \includegraphics[width=0.19\linewidth]{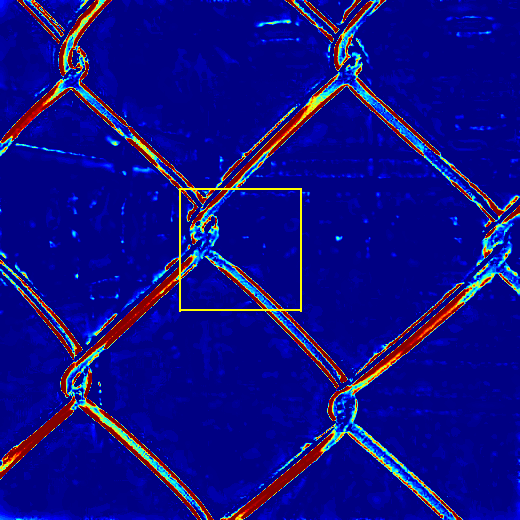}}
  \subfigure[GF]{\label{fig:subfig:f12}
    \includegraphics[width=0.19\linewidth]{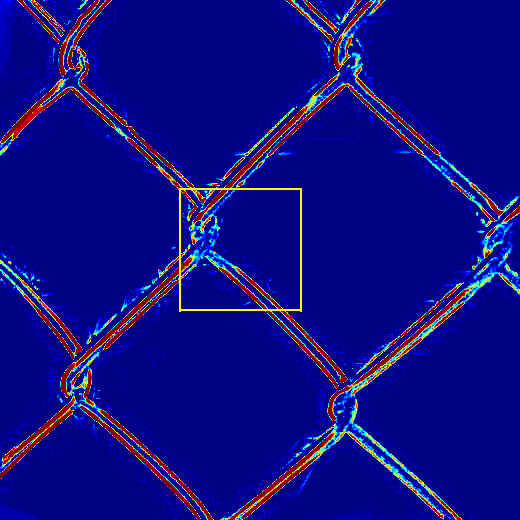}}
  \subfigure[DSIFT]{\label{fig:subfig:g12}
    \includegraphics[width=0.19\linewidth]{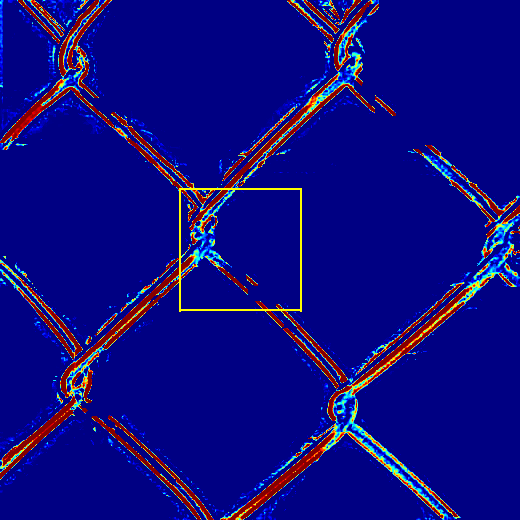}}
  \subfigure[CNN]{\label{fig:subfig:h12}
    \includegraphics[width=0.19\linewidth]{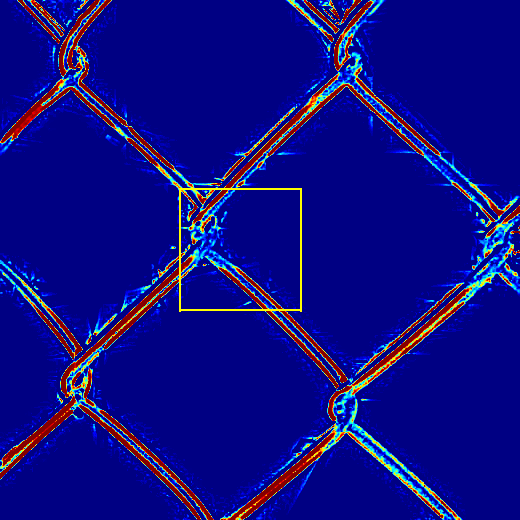}}
  \subfigure[The proposed method]{\label{fig:subfig:i12}
    \includegraphics[width=0.19\linewidth]{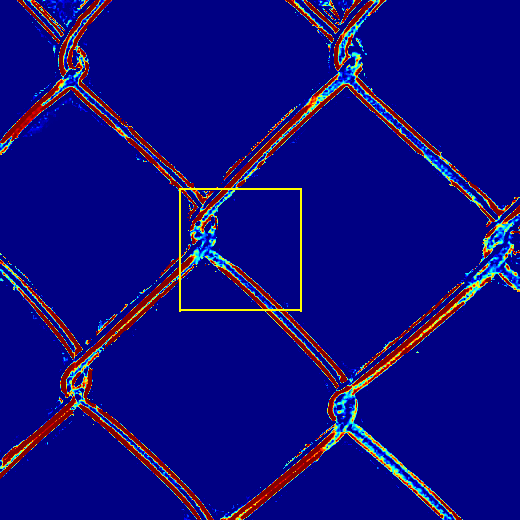}}
  \caption{The difference maps of different fusion methods with source image B.}
\label{fig:subfig:4} 
\end{figure*}

\subsection{Network Details and Training Settings}

In our implementation, typical Res56 networks are employed. The influences of network depth and the size of patches are tested. The performance will be improved if a deeper network or larger patches are used. Nonetheless, there is a trade-off between performance with time and memory cost. The chosen 56-layer networks with input patch size of 64*64*2, have produced satisfied results.

The same structure is applied to the initial fusion net and the boundary refinement net, and the training is carried out on different sample sets. The initial fusion net is trained on the full sample set, while the boundary refinement net is trained only on the samples that are near the FDB. Only the patches having at least 10 percent focused pixels and 10 percent defocused pixels is chosen.

As for network settings, the normal setups of ResNet on CIFAR-10 \cite{RN618} are used, since CIFAR-10 has similar input as our approach. The stochastic gradient descent (SGD) is applied to train the models, with softmax loss function. 15 percents of the training examples are randomly handout for verification. The initial learning rate is set to 0.001, and reduced after 80, 120, 160, 180 epochs. 200 epochs are used in total, and the batch size is 128. Therefore, there are about 15000 iterations during one single epoch in total.

\subsection{Comparison Settings}

We compare the proposed approach with 6 other multi-focus fusion methods, including NSCT \cite{ZHANG20091334}, SR \cite{RN35}, NSCT-SR \cite{LIU2015147}, GF \cite{RN80}, DSIFT \cite{RN43}, and CNN \cite{RN497}.

The comparison is conducted on 24 pairs of multi-focus images: 20 pairs from dataset "Lytro", the mostly mentioned dataset for image fusion; the others are "clock", "book", "lab" and "flower", which are commonly used too.

A few objective metrics for fusion image quality assessment are used to evaluate the results: normalized mutual information ($Q_{NMI}$) \cite{4610673}, gradient-based fusion performance ($Q_{G}$) \cite{RN103}, Yang's metric ($Q_{Y}$) \cite{YANG2008156}, Chen-Blum metric ($Q_{CB}$) \cite{CHEN20091421}. 

\subsection{Experimental Results and Analysis}

Fig. \ref{fig:subfig:3} shows the visual comparison on the "clock". Compared with other approaches, the proposed method performs well in general. Moreover, both the front clock's edge and the left side of number "8" in the behind clock are clear in our result, as shown in the enlarged squares.

Fig. \ref{fig:subfig:4} shows another example, "lytro-05". The difference maps ($DM$) is made to show the enhanced visual result: 
\begin{eqnarray}
DM\,= \alpha \cdot \left|\,ImgS_B - ImgF\,\right|.
\end{eqnarray}
Here $\alpha$ is set to 10, showing a more straightforward result. The difference maps are made into gray scale and the pseudo-color transformation is then applied. The difference map is expected to be red in the area where the source image is focused, and blue where the source image is blurred. Source image B is blurred on the iron grid and clear on the rest area. Compared with existing approaches, the difference map of our fusion result is distinct near the FDB. There is less incorrect division in the dark part of difference map, and the bright area is more continuous and clear than the others too.

The quantitative comparisons are shown in Table \ref{table11} and the results of proposed method without and with refinement are put in the last two columns. The larger evaluation means better fusion result for all the four quality metrics. The shown results are the average values over the 24 pairs of images, and the best average result of the compared methods is bold. The number of image pairs that one method beats all the other methods is shown in the parentheses. In addition, the without refinement version is ignored in the above comparisons.

The proposed method without refinement already outperforms existing methods, then the refinement improves the results, especially the $Q_{CB}$. It means that both the modification of the network structure and the refinement are effective. As can be seen, the proposed approach markedly outperforms the other fusion methods on all the four quality metrics. 

\section{Conclusions}

\begin{table*}[!ht]
\centering
\caption{The quantitative comparison of different fusion methods on 24 widely used image pairs. The results of the proposed method without/with refinement are shown in the last two columns. The best average result of the compared methods is in bold. The number of image pairs that one method beats all the other methods is shown in the parentheses. Particularly, the without refinement version is ignored in this comparison. }
\label{table11}
\begin{tabular}{|c|c|c|c|c|c|c|c|c|}
\hline
Metrics & NSCT &  SR & NSCT-SR  & GF  & DSIFT  &   CNN & \multicolumn{2}{|c|}{\tabincell{c}{Proposed Method \\without/with Refinement }}   \\
\hline
  $Q_{MI}$  & 0.9407 (0)  & 1.0807 (0) & 0.9651 (0) & 1.1001 (0) & 1.1902 (1) & 1.1561 (0) & \ \ \ 1.2000 \ \ \   & \textbf{1.2002} (23)  \\
\hline
  $Q_G $    & 0.6789 (0)  & 0.6944 (0) & 0.6823 (0) & 0.7104 (0) & 0.7162 (0) & 0.7155 (5) &  0.7187  & \textbf{0.7188} (18)  \\
\hline
  $Q_Y $    & 0.9549 (0)  & 0.9663 (0) & 0.9576 (0) & 0.9775 (1) & 0.9841 (2) & 0.9851 (6) &  0.9867  & \textbf{0.9871} (16)  \\
\hline
  $Q_{CB}$  & 0.7423 (0)  & 0.7641 (0) & 0.7499 (0) & 0.7848 (0) & 0.8005 (7) & 0.8000 (1) &  0.8028  & \textbf{0.8042} (16)  \\
\hline
\end{tabular}
\end{table*}

In this paper, we present a boundary aware multi-focus fusion approach base on deep neural networks. The proposed method utilizes residual networks, and a 2-channel model is applied to extract more useful information directly from the source images. Moreover, independent refinement networks are employed after the initial net, to deal with the different situations for area near and far away from the FDB, respectively. Furthermore, a new way to generate training samples is also proposed to better approximate the reality. Based on all these improvements, the proposed method obtains promising results and outperforms the state-of-the-art methods, both qualitatively and quantitatively.

\section{ACKNOWLEDGEMENT}
This work was supported by the National Natural Science Foundation of China under Grant No. 61771276.

\small
\bibliographystyle{IEEEbib}
\bibliography{icme2019template}

\end{document}